# Explainable Computer Vision Framework for Automated Pore Detection and Criticality Assessment in Additive Manufacturing

Akshansh Mishra[1,2], Rakesh Morisetty[3]

[1]School of Industrial and Information Engineering, Politecnico di Milano, Milan, Italy
[2]Computational Materials Research Group, AI Fab Lab, Uttar Pradesh, India
[3] Fastweb S.p.A., Milan, Italy

**Abstract:** Internal porosity remains a critical defect mode in additively manufactured components, compromising structural performance and limiting industrial adoption. Automated defect detection methods exist but lack interpretability, preventing engineers from understanding the physical basis of criticality predictions. This study presents an explainable computer vision framework for pore detection and criticality assessment in three-dimensional tomographic volumes. Sequential grayscale slices were reconstructed into volumetric datasets, and intensity-based thresholding with connected component analysis identified 500 individual pores. Each pore was characterized using geometric descriptors including size, aspect ratio, extent, and spatial position relative to the specimen boundary. A pore interaction network was constructed using percentile-based Euclidean distance criteria, yielding 24,950 inter-pore connections. Machine learning models predicted pore criticality scores from extracted features, and SHAP analysis quantified individual feature contributions. Results demonstrate that normalized surface distance dominates model predictions, contributing more than an order of magnitude greater importance than all other descriptors. Pore size provides minimal influence, while geometric parameters show negligible impact. The strong inverse relationship between surface proximity and criticality reveals boundary-driven failure mechanisms. This interpretable framework enables transparent defect assessment and provides actionable insights for process optimization and quality control in additive manufacturing.

## 1. Introduction

Additive manufacturing has transformed production capabilities across aerospace, biomedical, and automotive industries through layer-wise material deposition that enables complex geometries unattainable through conventional methods [1-5]. However, the same process attributes that provide design freedom also introduce microstructural defects, particularly internal porosity, which compromise mechanical integrity and fatigue life [6-9]. Pores nucleate from trapped gas, incomplete fusion, or powder bed irregularities during solidification, and their size, shape, and spatial distribution directly influence crack initiation and propagation under operational loads.

Current quality control relies heavily on computed tomography to visualize internal defects, yet the analysis remains largely manual or semiautomated with limited physical insight into which pores pose the greatest risk. Traditional machine learning models can predict failure modes from pore characteristics, but these models function as black boxes that offer predictions without mechanistic justification [10-15]. This opacity prevents engineers from understanding why certain pores are classified as critical, limiting confidence in automated assessment systems and hindering process optimization.

Explainable artificial intelligence offers a pathway to transparent defect analysis through techniques that quantify individual feature contributions to model predictions [16-18]. Recent applications of SHAP values in materials science have demonstrated the capacity to rank feature importance and reveal interaction effects, yet their application to three-dimensional pore networks in additive manufacturing remains unexplored. Understanding which geometric and spatial descriptors govern pore criticality could inform scanning strategies, process parameter selection, and post-processing decisions.

This work addresses the need for an interpretable framework that not only detects pores in tomographic volumes but also explains the physical basis for their predicted criticality scores, enabling data-driven quality assurance with mechanistic transparency.

## 2. Methodology

The input data consist of a three-dimensional tomographic image stack stored as sequential grayscale TIFF slices. Each slice represents a cross-section of the specimen [19], and stacking along the axial direction reconstructs the full volumetric dataset $I(z, y, x)$. All images were loaded and assembled into a 3D array while preserving original voxel intensities. Only high-intensity voxels were considered in subsequent analysis, corresponding to bright regions in the tomographic images in order to suppress non-relevant background information and isolate pore-related features. Pore candidates were identified using intensity-based thresholding computed using Equation 1.

$$M(z, y, x) = 1\{I(z, y, x) \geq I_{thr}\} \qquad (1)$$

Where $I_{thr} = 250$ was selected to capture bright pore regions. Connected-component labeling was applied to the binary mask to identify individual bright regions shown in Figure 1. The largest connected component was consistently found to correspond to the specimen boundary or imaging artifact and was therefore excluded from further analysis. Remaining connected components were classified as pores if their voxel count satisfied Equation 2.

$$2 < A_k < 0.01 A_{max} \qquad (2)$$

Where $A_k$ is the pore volume and $A_{max}$ is the size of the largest connected component.

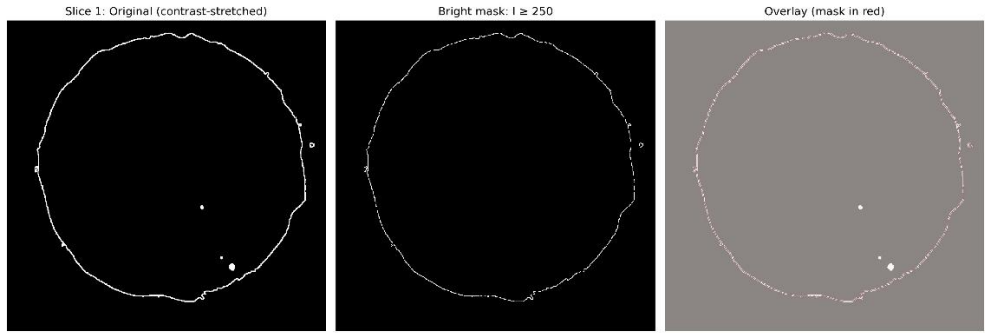

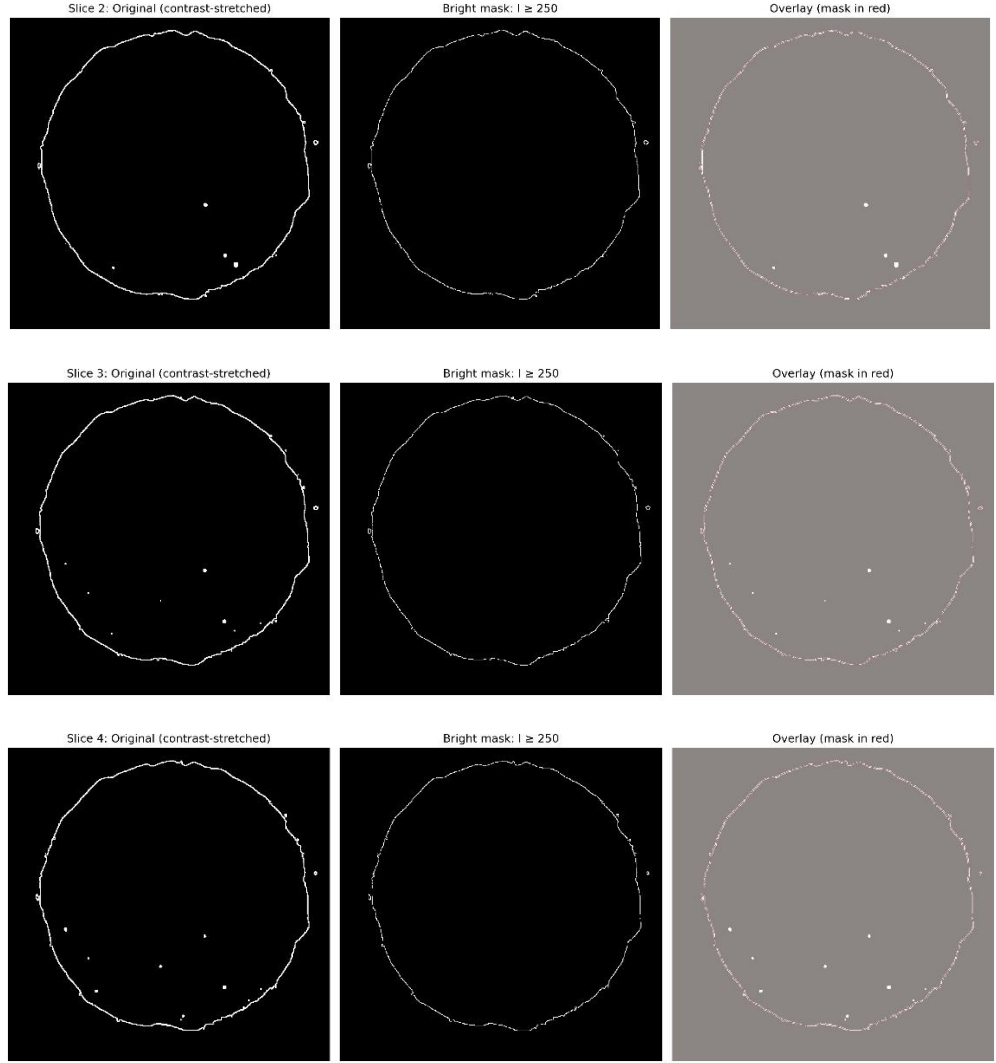

**Figure 1.** Visualization of pore segmentation on representative tomographic slices. For each slice, the left panel shows the contrast-stretched original grayscale image, the middle panel shows the high-intensity mask ($I \geq 250$) highlighting the dominant bright boundary, and the right panel shows the overlay of the bright mask on the original image (red). The figure illustrates that the high-intensity threshold primarily captures the specimen boundary, while interior pores exhibit lower intensities and require adaptive thresholding for reliable detection.

Each identified pore was characterized using a set of geometric and spatial descriptors after segmentation process. The centroid of each pore was computed to define its three-dimensional location within the specimen. Pore size was quantified as the total number of voxels comprising each pore. Additional shape descriptors, including aspect ratio and extent, were extracted to capture pore morphology. The axial position of each pore was defined by the centroid coordinate along the stacking direction. The minimum distance from each pore centroid to the specimen surface was computed and normalized by the specimen dimensions to account for boundary effects.

A pore interaction network was constructed by representing each pore as a node positioned at its centroid. Pairwise distances between pores were evaluated, and connections were established using a percentile-based distance criterion to ensure consistent network density. This approach avoids the need for an arbitrary global distance threshold and enables reproducible comparison across datasets. The resulting network encodes spatial proximity relationships between pores while remaining independent of pore size or orientation. Network size was limited to the largest pores to maintain computational tractability without altering the underlying spatial structure.

A supervised machine learning model was trained to predict a scalar criticality score for individual pores using the extracted descriptors as input features. The feature set includes both intrinsic pore characteristics and extrinsic spatial information. Model training and evaluation were performed using standard data partitioning to ensure robustness and to prevent overfitting. Predictions were generated at the pore level, allowing local structural features to be directly associated with model outputs. Explainable artificial intelligence techniques based on SHapley Additive exPlanations were employed to interpret the model predictions. Feature importance was quantified using mean absolute SHAP values, while beeswarm and dependency plots were used to analyze the directionality and variability of feature contributions. This analysis enabled identification of dominant controlling features and assessment of secondary effects without relying on model-specific assumptions.

### 3. Results and Discussion

The reconstructed three-dimensional pore network extracted from the segmented tomographic volume is shown in Figure 2. The multiple viewports represents the spatial organization and connectivity by identifying a total of 500 pores (nodes) and 24950 inter-pore connections (edges). Pores were identified as connected bright regions smaller than 1% of the dominant circular boundary area. The centroid position for each pore $k$ was computed using Equation 3.

$$c_k = \frac{1}{|\Omega_k|} \sum_{(z,y,x) \in \Omega_k} (z, y, x) \tag{3}$$

Figures 2a) and 2c) show that pores are distributed along the full specimen height (Z-direction), while remaining strongly localized near the outer radial boundary. Figure 2d) reveals an annular pore arrangement with a sparsely populated core that indicates a shell-dominated pore morphology rather than a uniformly porous volume.

Inter-pore connectivity was defined using the Euclidean distance between centroids computed using Equation 4.

$$d_{ij} = \|c_i - c_j\|_2 \tag{4}$$

Edges were constructed for all pore pairs satisfying Equation 5.

$$d_{ij} < d_{thr}, d_{thr} = Percentile_{20\%}(\{d_{ij} : i \neq j\})) \tag{5}$$

The total number of possible connections for $N = 500$ pores is computed using Equation 6.

$$\binom{N}{2} = \frac{N(N-1)}{2} = 124750 \tag{6}$$

The retained lowest 20% of the distance yields $0.2 \times 124750 = 24950$ edges which exactly matches the observed network density shown in Figure 2 (a-d). This confirms that the connectivity level is mathematically consistent and controlled.

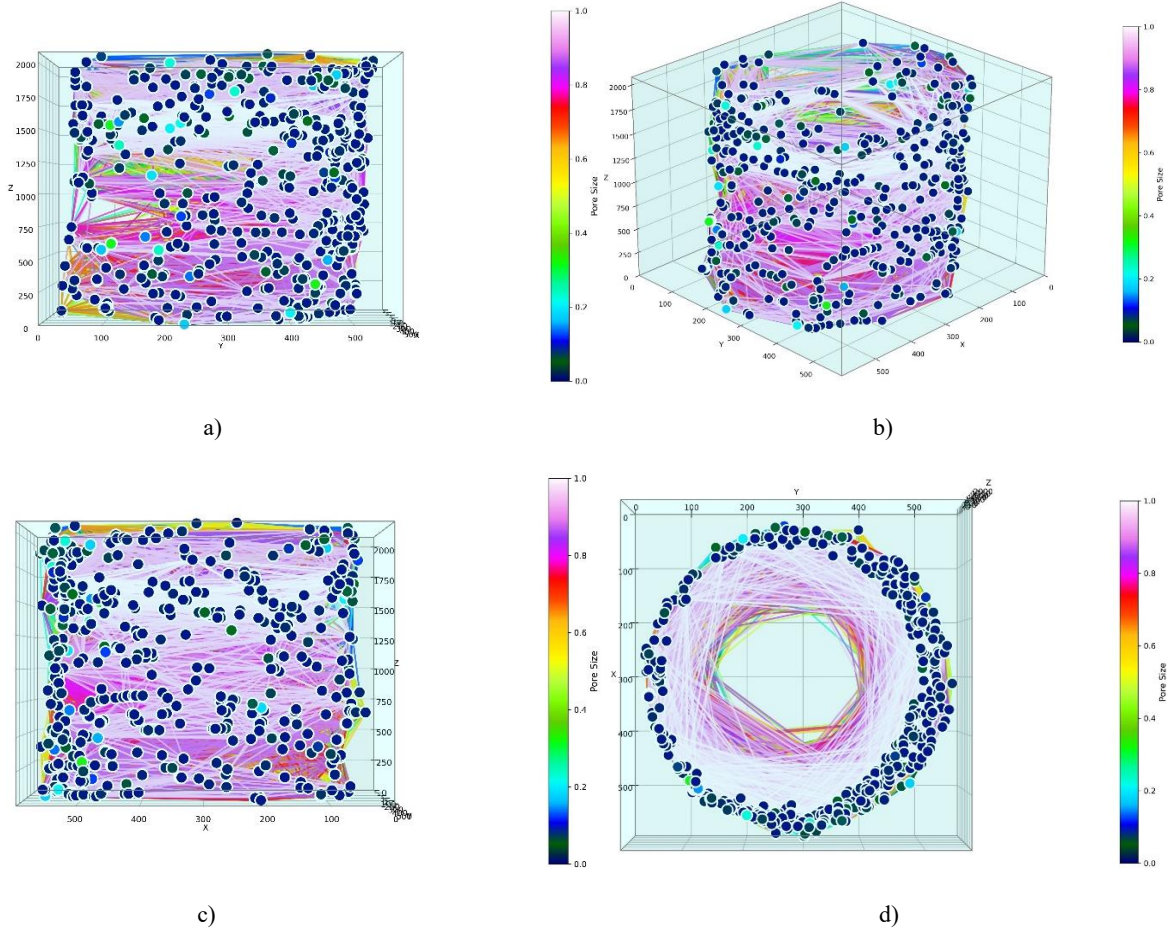

**Figure 2.** Three-dimensional pore network reconstructed from the segmented tomographic volume. The network consists of 500 pores (nodes) and 24,950 inter-pore connections (edges) defined using a percentile-based Euclidean distance criterion. Nodes correspond to pore centroids and are colored by normalized pore size, while edges represent spatial proximity between pores that show different viewing orientations: (a) front view, (b) isometric view, (c) side view, and (d) top view.

The SHAP-based feature importance ranking shown in Figure 3 demonstrates a dominance of surface distance in the determination of model response. The mean absolute SHAP value of surface distance exceeds that of all other features by more than an order of magnitude, indicating that the model predictions are primarily governed by pore proximity to the external boundary. It is observed that the pore size contributes only marginally, while geometric descriptors such as aspect ratio, extent, and axial ($Z$) position have negligible influence on the prediction outcome. This hierarchy implies that spatial location outweighs intrinsic pore morphology in the learned relationship. The model output $f(x)$ can be approximated using Equation 7.

$$f(x) \approx \Phi_{surface\ distance} + \epsilon \tag{7}$$

Where $\Phi$ denotes the SHAP contribution and $\epsilon$ aggregates all remaining features. The small magnitude of $\epsilon$ confirms that secondary descriptors do not significantly alter the prediction once surface distance is known.

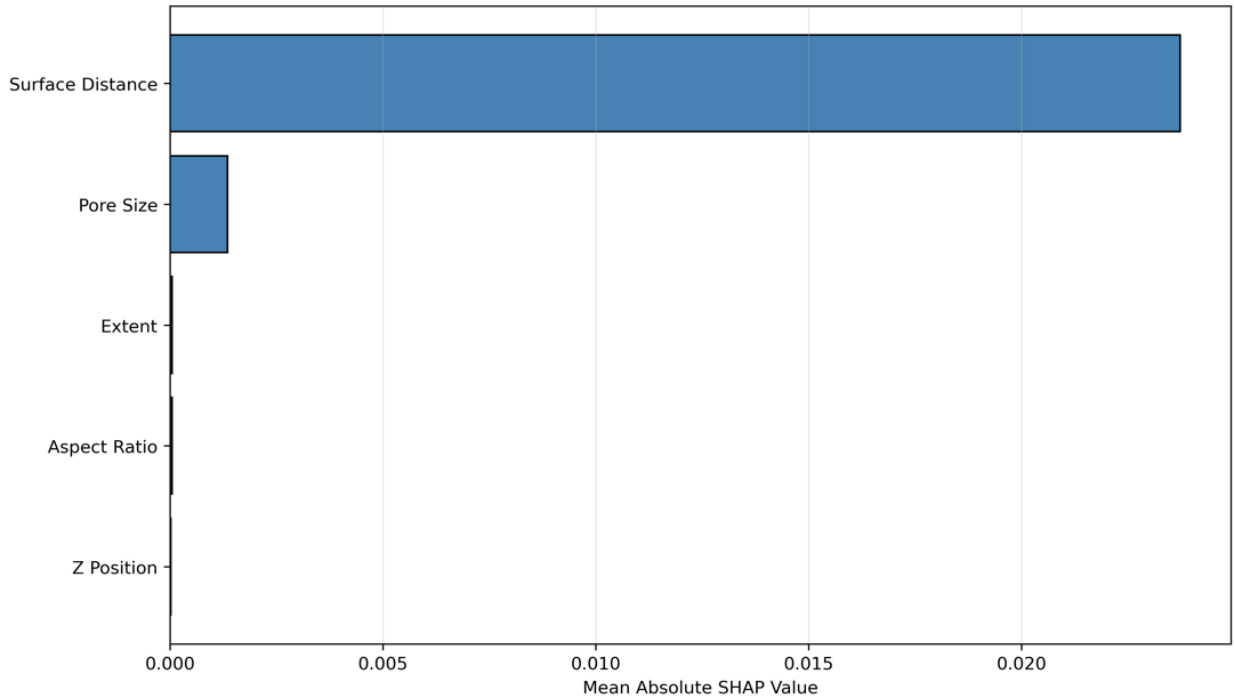

**Figure 3.** SHAP-based feature importance ranking showing the mean absolute SHAP values for all input features. Surface distance dominates the model prediction, contributing more than an order of magnitude higher importance than all other descriptors.

The SHAP beeswarm plot shown in Figure 4 reveals both the magnitude and directionality of feature contributions to the model output. Surface distance shows the widest SHAP value distribution, confirming it as the dominant feature controlling the prediction. High surface distance values (red points) are associated with negative SHAP values, while low surface distance values (blue points) contribute positively to the model output. This indicates a strong, monotonic inverse relationship between surface distance and the predicted response. Pore size exhibits a limited but interpretable influence. Larger pores (red) tend to shift the prediction slightly in the positive direction, whereas smaller pores (blue) have a weak negative or near-zero effect. The narrow spread of SHAP values indicates that pore size acts as a secondary modifier rather than a primary driver. Extent, aspect ratio, and axial ($Z$) position show SHAP values tightly clustered around zero with no clear color separation. This implies that these descriptors neither introduce strong nonlinear effects nor interact significantly with the dominant features. Their contributions remain negligible across their entire value ranges.

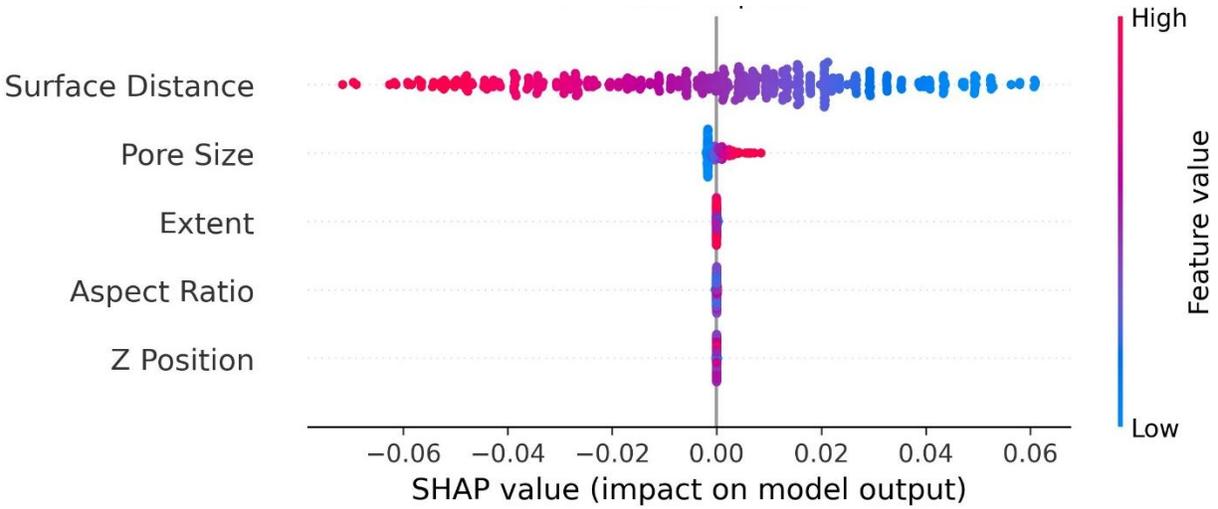

**Figure 4.** SHAP beeswarm plot illustrating the distribution, magnitude, and direction of feature contributions to the model output. Each point represents an individual sample, colored by feature value (blue: low, red: high).

The scatter plot shown in Figure 5 depicts the relationship between pore size and the predicted criticality score, with point color indicating the SHAP contribution of pore size. The distribution reveals no strong monotonic dependence between pore size and criticality across the majority of the data. Small pores (≈3–8 voxels) span nearly the full range of criticality values, indicating that pore size alone does not determine system criticality. This observation is consistent with the low mean SHAP importance previously observed for pore size. Larger pores (≥15 voxels) exhibit a slightly reduced spread in criticality and tend to cluster around intermediate-to-high values. The corresponding SHAP values become positive (yellow–red points), suggesting that pore size acts as a weak positive modifier rather than a primary driver of the model output. The absence of sharp trends or bifurcation points implies that critical behavior is not governed by pore size thresholds. Instead, pore size contributes additively and conditionally, likely interacting with spatial features such as surface distance.

The scatter plot shown in Figure 6 depicts a strong, nearly linear inverse relationship between normalized surface distance and the predicted criticality score. Pores located closer to the surface consistently exhibit higher criticality, while pores deeper within the material show progressively lower criticality values. The smooth monotonic trend and narrow scatter indicate that surface proximity is a deterministic control parameter rather than a weak statistical correlate. This observation explains the dominant SHAP importance previously identified for surface distance, as variations in this feature alone are sufficient to account for most of the variance in model predictions. The color gradient further reinforces this interpretation. High SHAP values (yellow–orange) correspond to small surface distances and positive contributions to criticality, whereas larger distances yield negative SHAP contributions (purple), reducing the predicted response. This confirms a stable and consistent directionality of influence across the entire feature range. Physically, this behavior suggests that the governing mechanism underlying criticality is boundary-driven, with surface-adjacent pores playing a disproportionately important role. Such behavior may arise from enhanced stress concentration, transport accessibility, or boundary-condition effects localized near the specimen surface.

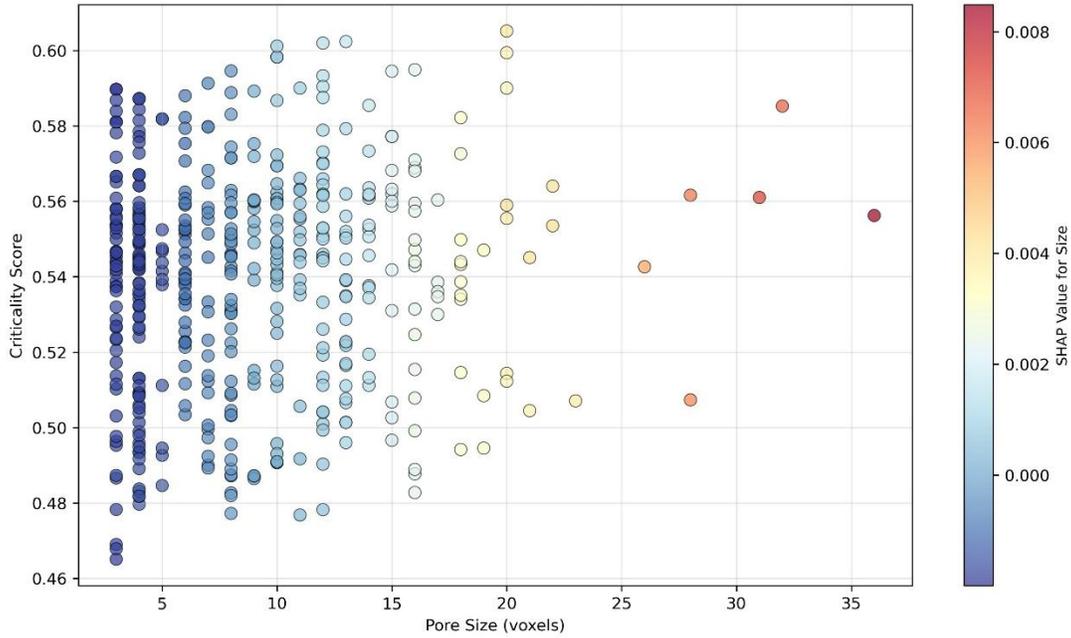

**Figure 5.** Relationship between pore size and predicted criticality score. Each point represents an individual pore, with color indicating the SHAP value associated with pore size. The broad scatter and lack of a strong monotonic trend indicate that pore size alone does not govern criticality. Larger pores exhibit a weak positive contribution to the prediction, acting as a secondary modifier rather than a primary controlling feature.

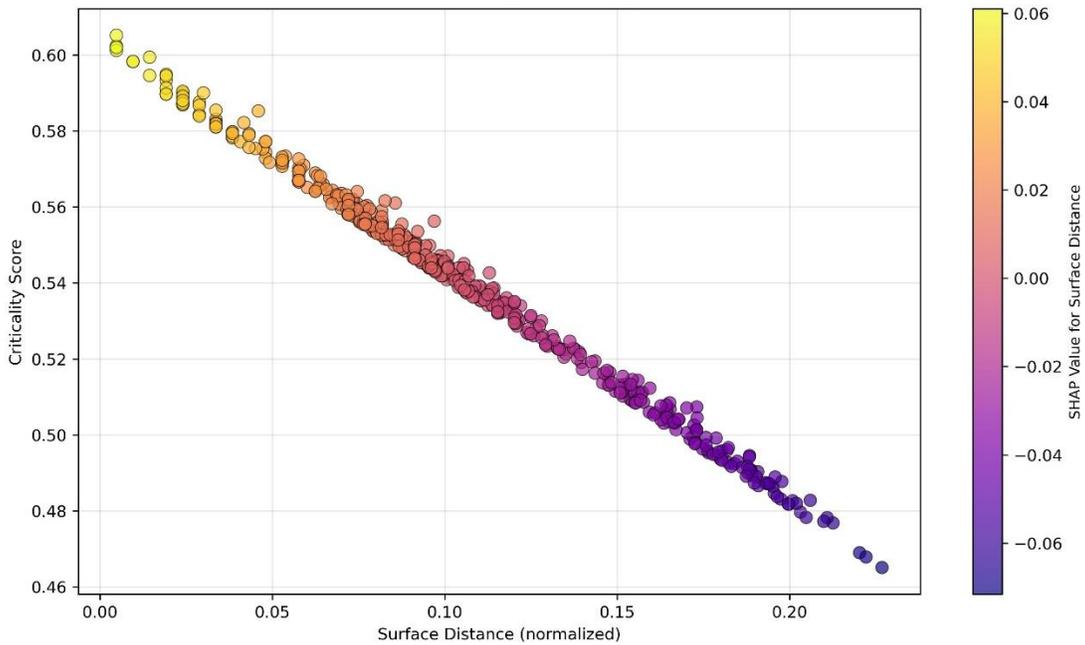

**Figure 6.** Relationship between normalized surface distance and predicted criticality score. Each point represents an individual pore, colored by the SHAP value associated with surface distance. A strong inverse, nearly linear trend is observed, indicating that pores closer to the surface exhibit higher criticality. The consistent SHAP color gradient confirms that surface proximity is the dominant and directionally stable contributor to the model prediction.

## 4. Conclusions

This study developed an explainable computer vision framework for automated pore detection and criticality assessment in additive manufacturing components using three-dimensional tomographic data. Analysis of 500 segmented pores and 24,950 spatial connections revealed that normalized surface distance dominates criticality predictions, exceeding all other feature contributions over an order of magnitude. The strong inverse relationship between surface proximity and predicted criticality indicates that boundary-adjacent pores pose disproportionate structural risk compared to interior defects.

Pore size demonstrated minimal predictive influence despite its conventional association with failure modes, while geometric descriptors including aspect ratio and extent contributed negligibly to model outputs. SHAP analysis quantified these relationships transparently, confirming that spatial location rather than intrinsic pore morphology governs criticality in the examined specimens. This mechanistic insight suggests that scanning strategies and process controls should prioritize near-surface defect mitigation over bulk porosity reduction.

The interpretable framework enables quality assurance decisions grounded in physical understanding rather than opaque algorithmic predictions. Future work should extend this methodology across different alloy systems, manufacturing processes, and loading conditions to validate the generalizability of surface-driven criticality mechanisms and develop adaptive inspection protocols for additive manufacturing quality control.